\documentclass{article}

\usepackage[final]{corl_2020} 
\usepackage{stdlib}

\title{
  Robot Action Selection Learning via Layered \\Dimension Informed Program Synthesis
}

\author{
Jarrett Holtz,
Arjun Guha, and
Joydeep Biswas
\\
University of Texas at Austin, and University of Massachusetts Amherst \\
\{jaholtz,joydeepb\}@cs.utexas.edu, arjun@cs.umass.edu
}

\begin{document}

\maketitle

\begin{abstract}
 Action selection policies (ASPs), used to compose low-level robot skills into
complex high-level tasks are commonly represented as neural networks (NNs) in
the state of the art.
Such a paradigm, while very effective, suffers from a few key problems: 1)
NNs are opaque to the user and hence not amenable to verification, 2) they
require significant amounts of training data, and 3) they are hard to repair
when the domain changes.
We present two key insights about ASPs for robotics. First, ASPs need to
reason about physically meaningful quantities derived from the state of the
world, and second, there exists a layered structure for composing these
policies.
Leveraging these insights, we introduce layered dimension-informed program
synthesis (LDIPS) -- by reasoning about the physical dimensions of state
variables, and dimensional constraints on operators, LDIPS directly synthesizes
ASPs in a human-interpretable domain-specific language that is amenable to
program repair. We present empirical results to demonstrate that LDIPS 1) can
synthesize effective ASPs for robot soccer and autonomous driving domains,
2) enables tractable synthesis for robot action selection policies not possible
with state of the art synthesis techniques,
3) requires two orders of magnitude fewer training examples than a comparable NN
representation, and 4) can repair the synthesized ASPs with only a small
number of corrections when transferring from simulation to real robots.

\end{abstract}


\section{Introduction}
  End-users of service mobile robots want
the ability to teach their robots how to perform novel tasks, by
composing known low-level skills into high-level behaviors based on
demonstrations and user preferences.
Learning from Demonstration (LfD)~\cite{lfdSurvey}, and Inverse Reinforcement Learning
(IRL)~\cite{ziebartInverse} have been applied to solve this problem, to great success in
several domains, including furniture assembly~\cite{nieukumFurniture}, object pick-and-place
~\cite{levineDemo}, and surgery~\cite{padoySurgery, kellerSurgery}.
A key driving factor for these
successes has been the use of Neural Networks (NNs) to learn
the action selection policy (ASP) directly~\cite{schulman2015trust,schulman2017proximal}, or the value function from
which the policy is derived~\cite{tamar2016value}.
Unfortunately, despite their success at
representing and learning policies, LfD using NNs suffers from the following
well-known problems:
1) they are extremely data-intensive, and need a variety of demonstrations
before a meaningful policy can be learned~\cite{deepLearningRobots};
2) they are opaque to the user, making it hard to understand \emph{why} they
do things in specific ways or to verify them~\cite{manuelaExplanation};
3) they are quite brittle, and very hard to repair when parameters of the
problem change, or when moving from
simulation to real robots~\cite{closingSimToReal}.

We present the following observations about ASPs independent of their
representation:
1) The input states to a policy consist of
\emph{physically meaningful quantities}, \eg{} velocities, distances, and angles.
2) The structure of a policy has \emph{distinct levels of abstraction},
including computing relevant features from the state, composing several
decision-making criteria, and making decisions based on task- and domain-
specific parameters.
3) A well-structured policy is easy to repair in terms of only the parameters
that determine the decision boundaries, when the domain changes.

Based on these insights we build on \textit{program synthesis} as a means
to address the shortcomings of neural approaches.
Program synthesis seeks to automatically find a program
in an underlying programming language that satisfies some user specification
\cite{gulwani2017program}.
Synthesis directly addresses these concerns by learning policies as
human-readable programs, that
are amenable to program repair, and can by do so
with only a small number of demonstrations as a specification.
However, due to two major limitations,
existing state of the art synthesis approaches are not sufficient for learning
robot programs. First, these approaches are not designed to handle non-linear
real arithmetic, vector operations, or dimensioned quantities, all commonly
found in robot programs. Second, synthesis techniques are largely limited by
their ability to scale with the search space of potential programs, such that
ASP synthesis is intractable for existing approaches.

To address these limitations and apply synthesis to solving the LfD problem
we propose \emph{Layered Dimension-Informed Program Synthesis} (LDIPS).
We introduce a domain-specific language (DSL) for
representing ASPs where a type system keeps track of the physical
dimensions of expressions, and enforces dimensional constraints on mathematical
operations. These dimensional constraints limit the search space of the program,
greatly
improving the scalability of the approach and the performance of the resulting
policies.
The DSL structures ASPs into decision-making criteria for each
possible action, where the criteria are repairable parameters, and the
expressions used are derived from the state variables. The inputs to LDIPS are a
set of sparse demonstrations and an optional \emph{incomplete ASP},
that encodes as much structure as the programmer may have
about the problem. LDIPS then fills in
the blanks of the incomplete ASP using syntax-guided
synthesis~\cite{syntaxGuided} with dimension-informed expression and operator
pruning. The result of LDIPS
is a fully instantiated ASP, composed of synthesized features, conditionals,
and parameters.

We present empirical results of applying LDIPS to robot soccer and
autonomous driving, showing that it is capable of generating ASPs that
are comparable in performance to expert-written ASPs that performed well
in a (ommitted for double-blind review) competition. We evaluate experimentally
the effect of dimensional constraints on the performance of the policy and
the number of candidate programs considered.
We further show that LDIPS is capable of synthesizing such
ASPs with two orders of magnitude fewer examples than an NN representation.
Finally, we show that LDIPS can synthesize ASPs in simulation, and given only a
few corrections, can repair the ASPs so that they perform
almost as well on the real robots as they did in simulation.

\section{Related Work}
  \seclabel{related}
  The problem of constructing ASPs from human demonstrations has been extensively
studied in the LfD, and inverse reinforcement learning (IRL) settings~\cite{reinSurvey,deepLearningRobots,lfdSurvey}.
In this section, we focus on
\begin{paraenum}
  \item alternative
    approaches to overcome data efficiency, domain transfer, and
    interpretability problems;
  \item concurrent advances in program synthesis;
  \item recent work on symbolic learning
    similar to our approach;
  \item synthesis and formal methods applied to robotics.
\end{paraenum}.
We conclude with a summary of the our contributions compared to the
state of the art.

The field of transfer learning attempts to address generalization and
improve learning rates to reduce data requirements~\cite{stoneTransferSurvey}.
Model-based RL can also reduce the data requirements on real robots, such
as by using dynamic models to guide simulation~\cite{dudekModelBased}.
Other work addresses the problem of generalizing learning by
incorporating corrective demonstration when errors are encountered
during deployment~\cite{thomazCorrectiveDemo}.
Approaches to solving the Sim-to-Real problem
have modified the training process and adapted simulations
\cite{closingSimToReal}, or utilized progressive nets to
transfer features \cite{simToRealProg}.
Recent work on interpreting policies has focused on finding interpretable
representations of NN policies, such as with Abstracted Policy
Graphs~\cite{manuelaExplanation},
or by utilizing program synthesis to mimic the NN policy~\cite{swaratInterp}.

SyGuS is a broad field of synthesis techniques that have been applied in
many domains~\cite{syntaxGuided}.
The primary challenge of SyGuS is scalability, and there are many approaches
for guiding the synthesis in order to tractably find the best programs.
A common method for guiding synthesis is the use of \emph{sketches}, where
a sketch is a partial program with some \emph{holes} left to be filled in
via synthesis~\cite{combinatorialSketching}.
Another approach is to quickly rule out portions of the program space that
can be identified as incorrect or redundant, such as by identifying
equivalent programs given examples \cite{exampleTransit},
by learning to recognize properties that make candidate programs invalid~\cite{isilConflictDriven},
or by using type information to identify
promising programs~\cite{typeExample}. A similar approach is to consider sets
of programs at once, such as by using a special data structure
for string manipulation expressions \cite{gulwaniStrings}, or by using
SMT alongside sketches to rule out multiple programs simultaneously~\cite{swaratComponent}.

Recent symbolic learning approaches have sought to combine
synthesis and deep learning by leveraging NNs for
sketch generation \cite{swaratNeuralSketch, solarSketchLearn}, by guiding the search
using neural models \cite{gulwaniNeuralGuided}, or by leveraging purely
statistical models to generate programs \cite{robustFill}.
Alternatively, synthesis has been used to guide learning, as in work
that composes neural perception and symbolic program execution
to jointly learn visual concepts, words, and semantic parsing of questions
\cite{maoNeurSymbo}. While symbolic learning leveraging program synthesis
produces interpretable ASPs in restricted program spaces, these
approaches often still
require large amounts of data.

State-of-the-art work for synthesis in robotics focuses on three primary
areas.
The most related work uses SMT-based parameter repair alongside human corrections
for adjusting transition functions in robot behaviors \cite{holtz2018interactive}.
Similar work utilizes SyGuS as part of a symbolic learning approach to interpret
NN policies as PID controllers for autonomous driving
\cite{swaratInterp, swaratImitation}.
A different, but more common synthesis strategy in robotics is reactive synthesis.
Reactive synthesis produces correct-by-construction policies based on
Linear Temporal Logic specifications of behavior by generating policies as
automata without relying on a syntax~\cite{kressSurvey, kressPlanning, topcuReactive, salty}.

In this work, we present an LfD approach that addresses
data-efficiency, verifiability, and repairability concerns by utilizing SyGuS,
without any NN components.
\technique{} builds on past SyGuS techniques by introducing
dimensional-constraints. While past work in the programming languages community
has leveraged types for synthesis~\cite{typeExample}, to the best of
our knowledge none has incorporated dimensional analysis. Further, \technique{}
extends prior approaches by
supporting non-linear real arithmetic, such as trigonmetric functions, as well
as vector algebra.


\section{Synthesis for Action Selection}
\seclabel{technique}

\begin{figure}[t]

\begin{subfigure}[b]{0.4\textwidth}
    \centering
\includegraphics[width=0.80\textwidth]{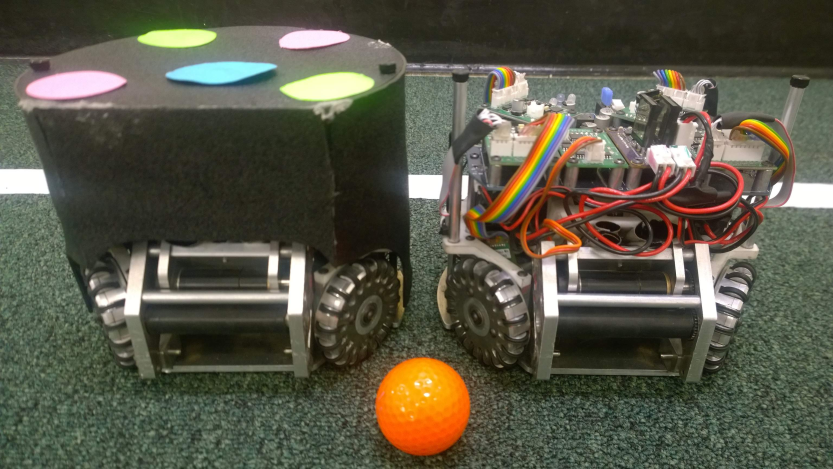}
\caption{RoboCup SSL robot that runs the ASP.}
\label{robot-pic}
\end{subfigure}
~
\begin{subfigure}[b]{0.5\textwidth}
\scriptsize

\fbox{\zdemo{Pre}{Post}{p_r,v_r,p_b,v_b}}

\zdemo{Goto}{Kick}{\langle -.1, .05\rangle,\langle 0,.1\rangle,\langle 0, 0\rangle,\langle 0, 0\rangle}

\zdemo{Goto}{Kick}{\langle -.1, .05\rangle,\langle 0,.1\rangle,\langle 0, 0\rangle, \langle 0, 0\rangle}

\zdemo{Inter}{Kick}{\langle -.1, .049\rangle,  \langle 0,.5\rangle,  \langle 0, 0\rangle,   \langle 0, .35\rangle}

\zdemo{Inter}{Inter}{\langle -.1, .49\rangle, \langle 0,.5\rangle, \langle 0, 0\rangle,  \langle 0, .651\rangle}

\zdemo{Inter}{Inter}{\langle -.1, .05\rangle,\langle 0,.5\rangle,\langle 0, 0\rangle,\langle 0, .35\rangle,}

\zdemo{Goto}{Inter}{\langle 0, 0\rangle,\langle 0,0\rangle,\langle .5, .5\rangle,\langle 0, -.201\rangle}

\zdemo{Goto}{Inter}{\langle 0, 0\rangle,\langle 0,0\rangle,\langle .5, .5\rangle,\langle .201, 0\rangle}

\zdemo{Goto}{Goto}{\langle 0, 0\rangle,\langle 0,0\rangle,\langle .5, .5\rangle,\langle -.1, .1\rangle}
\caption{Demonstrations.}
\label{example-demonstrations}
\end{subfigure}
~
\begin{subfigure}{0.5\textwidth}
\begin{lstlisting}[language=dipsexample]
if ($\startAction$==Kick || ($|p_r-p_b| < \hh{?\param}$ @\label{line:paramHole}@
    $\!\&\&\ |v_r-v_b| < \hh{?\param}$)): Kick @\label{line:paramHole2}@
elif ($\hh{?\expr > ?\param}$): Inter
else: Goto
\end{lstlisting}

\caption{Incomplete ASP provided by a user.}
\label{incomplete-asp-example}

\end{subfigure}
~
\begin{subfigure}{0.45\textwidth}
\begin{lstlisting}[language=dipsexample]
if ($\startAction$==Kick || ($|p_r-p_b| < \hh{150}$
    $\!\&\&\ |v_r-v_b| < \hh{100}$)): Kick
elif ($\hh{|v_b|} > \hh{100}$): Inter
else: Goto
\end{lstlisting}

\caption{LDIPS-completed ASP from user sketch.}
\label{complete-example}

\end{subfigure}
~
\begin{subfigure}[b]{0.5\textwidth}
\begin{lstlisting}[language=dipsexample]
if ($\startAction$==GoTo && $\hh{?\pred}$): Kick
else if ($\startAction$==Inter && $\hh{?\pred}$): Kick
else if ($\startAction$==Kick && $\hh{?\pred}$): Kick
else if ($\startAction$==GoTo && $?\hh{\pred}$): Inter
else if ($\startAction$==Inter && $?\hh{\pred}$): Inter
else if ($\startAction$==Kick && $\hh{?\pred}$): Inter
else: return GoTo
\end{lstlisting}

\caption{Incomplete ASP generated by \technique{}.}
\label{l3-incomplete-asp-example}

\end{subfigure}
~
\begin{subfigure}[b]{0.5\textwidth}
\begin{lstlisting}[language=dipsexample]
if ($\startAction$==Goto && $\hh{|p_r-p_b| < 150}$
    $\!\hh{\&\&\ |v_r-v_b| < 100}$): Kick
elif ($\startAction$==Inter &&
    $\hh{|v_r-v_b| < 150}$ $\! \hh{\&\&\ |p_r-p_b|  < 150}$): Kick
elif ($\startAction$==Goto && $\hh{|v_b| > 200}$): Inter
elif ($\startAction$==Inter && $\hh{|v_b| > 100}$): Inter
elif ($\startAction$==Kick && $\hh{|v_b| < 100}$): Kick
else: Goto
\end{lstlisting}

  \caption{LDIPS-completed ASP from \technique{} sketch.}
\label{l3-complete-example}

\end{subfigure}

\caption{Using the SSL robot (a), the user demonstrates an expected behavior.
Each demonstration is a transition from one action to another given the
position and velocity of the robot and ball (b). Then, the user may write
an incomplete sketch (c), that has blanks for \technique{} to fill out
(highlighted in blue). \technique{} then uses the demonstration to fill
in the blanks in the incomplete ASP (d). If the user provides no
sketch, then \technique{} generates a sketch from the demonstrations (e),
and fills in the blanks to form a complete program (f).}
\vspace{-1em}
\end{figure}

This section presents LDIPS, using our RoboCup soccer-playing robot as an
example (\cref{robot-pic}). We consider the problem of learning an action
selection policy (ASP) that directs our robot to intercept a moving ball and
kick it towards the goal. An ASP for this task employs three low-level
actions ($a$) to go to the ball (\texttt{Goto}), intercept it (\texttt{Inter}), and
kick it toward the goal (\texttt{Kick}). The robot runs the ASP
repeatedly, several times per second, and uses it to transition from one action
to another, based on the observed the position and velocity of the ball
($p_b,v_b$) and robot ($p_r,v_r$). Formally, an ASP for this problem
is a function that maps a previous action and a current world state to a next
action: $a \times \world \rightarrow a$. The world state definition is
domain-dependent: for robot soccer, it consists of the locations and velocities
of the ball and the robot
($\world = \langle p_b, v_b, p_r, v_r \rangle$).

An ASP can be decomposed into three logical layers:
1)~expressions that compute features (e.g., the distance to the ball, or its
velocity relative to the robot); 2) the construction of decision logic based on
feature expressions (e.g., the features needed to determine whether to kick
or follow the ball); and 3) the parameters that determine the decision
boundaries (e.g., the dimensions of the ball and robot determines the distance
at which a kick will succeed).

Given only a sequence of demonstrations, LDIPS can synthesize an ASP encoded as
a structured program. For example, \cref{example-demonstrations} shows a set of
nine demonstrations, where each is a transition from one action to another,
given a set of observations. Given these demonstrations, LDIPS generates an ASP
in three steps. 1)~It generates a sequence of \textbf{\texttt{if-then-else}}
statements that test the current action ($a_s$) and return a new action
(\cref{l3-incomplete-asp-example}). However, this is an \emph{incomplete ASP},
that has blank expressions ($?\expr$), and blank parameters ($?\param$).
2)~LDIPS uses bounded
program enumeration to generate candidate features. However, these features
have blank parameters for decision boundaries. 3)~LDIPS uses an SMT solver to
find parameter values that are consistent with demonstrations. If the currently
generated set of features is inadequate, then LIDPS will not
find parameter values. In that case, the algorithm will return to step (2) to
generate new features. Eventually, the result is a complete ASP that
we can run on the robot (\cref{l3-complete-example}).
Compared to other LfD approaches, a unique feature of LDIPS is that it can
also synthesize parts of an ASP with varying amounts of guidance.
For example, in addition to the demonstrations, the
user may also provide an incomplete ASP.
For example, the user can write the ASP shown in
\cref{incomplete-asp-example}, which has several blank parameters
($\hh{?x_p}$), e.g., to determine the maximum distance at which a \texttt{Kick}
will succeed. It also has blank expressions ($\hh{?e}$) and predicates
($\hh{?b}$), e.g., for the conditions under which the robot should kick a
moving ball. Given this incomplete ASP, LDIPS will produce a completed
executable ASP that preserves the non-blank portions of the incomplete ASP
(\cref{complete-example}).

\subsection{A Language for (Incomplete) Action Selection Policies}
\seclabel{grammar}

\begin{figure}
\scriptsize
\begin{subfigure}{0.65\textwidth}
\lstset{language=model}
\begin{minipage}{0.52\textwidth}
\(
\begin{array}{@{}l@{}c@{\;}l@{\;}l}
\multicolumn{3}{@{}l}{\textbf{Types and Dimensions}} \\
\env  & ::=  & [e:\type]      & \textrm{Type Environment} \\
\dims & ::=  & \dimension{}   & \textrm{Length, Time, Mass}  \\
      & \mid & [0, 0, 0]      & \textrm{Dimensionless} \\
\type & ::=  & \textrm{bool}  & \textrm{Boolean}  \\
      & \mid & \dims          & \textrm{Scalar with dimension \dims} \\
      & \mid & \vect{(\dims)} & \textrm{Vector with \dims-elements} \\[0.4em]
\multicolumn{3}{@{}l}{\textbf{Actions}} \\
\action & ::=  & a & \textrm{Domain-specific action} \\
        & \mid & \startAction & \textrm{Current action} \\
\multicolumn{3}{@{}l}{\textbf{Constant Values}} \\
\thresh & ::= &~\param : \dims & \textrm{Threshold Value} \\
        & \mid & ?\param : \dims & \textrm{Blank Parameter} \\
\end{array}
\)
\end{minipage}
~
\begin{minipage}{0.4\textwidth}
    \(\begin{array}{@{}r@{\;}c@{\;}l@{\;}l}
      \multicolumn{4}{@{}l}{\textbf{Action Selection Policy}} \\
      \pol & ::= & \texttt{return}~\action \\
      & \mid &
        \multicolumn{2}{@{}l}{\texttt{if (\pred): $\pol_1$ else $\pol_2$}} \\
      %
      \multicolumn{4}{@{}l}{\textbf{Predicates}} \\
      \pred & ::=  & \texttt{$\tru$~| $\fls$} \\
      & \mid & \action_1~\texttt{==}~\action_2
      & \hspace{0pt} \\
      & \mid & \expr~\texttt{>}~\thresh \\
      & \mid & \expr~\texttt{<}~\thresh \\
      & \mid & \pred_1~\&\&~\pred_2 \\
      & \mid & \pred_1~||~\pred_2 \\
      & \mid & ?\pred : \textrm{bool} & \textrm{Blank Predicate} \\
      \multicolumn{4}{@{}l}{\textbf{Expressions}} \\
      \expr & ::=  & \inputVar:\type & \textrm{Input Variable}\\
      & \mid & \const:\type & \textrm{Constant}\\
      & \mid & \multicolumn{2}{l}
      {\hspace{-6pt} \uOp{\expr}} \\
      & \mid & \multicolumn{2}{l}
      {\hspace{-6pt} \bOp{\expr_1}{\expr_2}} \\[.4em]
      & \mid & ?\expr:\type & \textrm{Blank Expression} \\[.4em]
    \end{array}\)
  \end{minipage}

\caption{The language of action selection policies.}
\label{syntax}
\end{subfigure}
~\vrule~
\begin{subfigure}{0.28\textwidth}
\scriptsize
\lstset{language=model}
\begin{minipage}[t!]{0.27\textwidth}
  \(\begin{array}{@{}c@{\;}c@{\;}l}
    \multicolumn{3}{@{}l}{\textbf{Unary Operators}} \\
    \mathit{op}_1 & ::= &
    \texttt{abs}
    \mid \texttt{sin}
    \mid \texttt{norm}
    \mid \cdots \\[.4em]
    \multicolumn{3}{@{}l}{\textbf{Binary Operators}} \\
    \mathit{op}_2 & ::= &
    \texttt{+}
    \mid \texttt{-}
    \mid \texttt{*}
    \mid \texttt{dist}
    \mid \cdots \\[.4em]
    \multicolumn{3}{@{}l}{\textbf{Skills}} \\
    \action & ::= &

      \mid \texttt{kick}
      \mid \texttt{cruise}
      \mid \cdots \\[.4em]
    \multicolumn{3}{@{}l}{\textbf{Types and Dimensions for Operations}} \\
    \textrm{abs} & : & \dims \rightarrow  \dims  \\
    \textrm{norm} & : & \vect(\dims) \rightarrow \dims  \\
    \textrm{sin} & : & [0,0,0] \rightarrow [0,0,0]  \\
    + & : & \dims \times \dims \rightarrow \dims  \\
    + & : & \vect(\dims) \times \vect(\dims) \rightarrow \vect(\dims)  \\
    * & : & \dims_1 \times \dims_2 \rightarrow \dims_1 + \dims_2  \\
    * & : & \dims_1 \times \vect(\dims_2) \rightarrow \vect(\dims_1 + \dims_2)  \\
    / & : & \dims_1 \times \dims_2 \rightarrow \dims_1 - \dims_2  \\
    / & : & \vect(\dims_1) \times \dims_2 \rightarrow \vect(\dims_1 - \dims_2)  \\
  \end{array}\)
\end{minipage}
\caption{The RoboCup domain.}
\label{robocup-domain}
\vspace{-1em}
\end{subfigure}

\caption{We write action selection policies in a simple, structured language
(\cref{syntax}). The language it supports several kinds of \emph{blanks}
($?b$, $?x_p$ $?e$), that LDIPS fills in. Every ASP relies on a set of
primtive actions and operators that vary across problem domains.
For example, \cref{robocup-domain} shows the actions and operators of
RoboCup ASPs. When LDIPS synthesizes an ASP, it uses the
domain definition to constrain the search space.}
\vspace{-1.5em}
\end{figure}

\Cref{syntax} presents a context-free grammar for the language of ASPs. In this
language, a policy ($\pol$) is a sequence of nested conditionals that return
the next action ($\action$). Every condition is a predicate ($\pred$) that
compares feature expressions ($\expr$) to threshold parameters ($\thresh$). A
feature expression can refer to input variables ($x_y$) and the value of the
last action ($a_s$). An \emph{incomplete ASP} may have blank expressions
($?e$), predicates ($?b$), or parameters ($?x_p$). The output of LDIPS is a
complete ASP with all blanks filled in. At various points in \technique{}
we will need to evaluate programs in this syntax with respect to a world
state, to accomplish this we employ a function \texttt{Eval($\pol, \world$)}.

Different problem domains require different sets of primitive actions and
operators. Thus for generality, LDIPS is agnostic
to the collection of actions and operators required. Instead, we instantiate
LDIPS for different domains by specifying the collection of actions ($\action$),
unary operators ($\mathit{op}_1$), and binary operators ($\mathit{op}_2$)
that are relevant to ASPs for that domain. For example, \cref{robocup-domain}
shows the actions and operators of the RoboCup domain.

The specification of every operator includes the types and dimensions of its
operands and result. In \secref{L2}, we see how LDIPS uses both types and dimensions
to constrain its search space significantly.
LDIPS supports real-valued scalars, vectors, and booleans with specific
dimensions.
Dimensional
analysis involves tracking base physical quantities
as calculations are performed, such that both the space of legal operations
is constrained, and the dimensionality of the result is well-defined. Quantities
can only be compared, added, or subtracted when they are commensurable,
but they may be multiplied or divided even when
they are incommensurable.
We extend the types $\type$ of our language with dimensions by defining
the dimension $\dims$ as the vector of dimensional
exponents $[n_1, n_2, n_3]$, corresponding to Length, Time, and Mass.
As an example, consider a quantity $a$:$t$,
if $a$ represents length in meters, then $t=[1,0,0]$,
and if $a$ represents a velocity vector with
dimensionality is $\textrm{Length} / \textrm{Time}$, then
$t = \vect{([1,-1,0])}$.
Further, we extend the type signature of operations to
include dimensional constraints that refine their domains and
describe the resulting dimensions in terms of the
input dimensions. The type signatures of operations, $\inputVar$, and $\const$
are represented in a type environment $\env$ that maps from expressions to types.



\subsection{\technique{-L1} : Parameter Synthesis}
\seclabel{L1}
\begin{wrapfigure}[10]{r}{0.35\textwidth}
  \vspace{-1.7em}
\begin{lstlisting}[language=Python, mathescape=true,numbers=none,
                   basicstyle=\scriptsize\ttfamily, escapechar=!]
!\(\begin{array}{@{}l}
  \textrm{\textbf{L1}} :
      \worlds \times \worlds \times \pred \rightarrow \pred~||~\textrm{UNSAT}
\end{array}\)!
L1($\posE$, $\negE$, $\pred$):
  !$\vec{?\param} = \textrm{ParamHoles}(\pred)$!
  !\(\begin{array}{@{}l@{}l@{}l}
    \phi & = & \exists \vec{?\param} \\
    & \mid &
    (\forall \world \in \posE~.~\textrm{PartialEval}(\pred, \world))\land \\
    & &
    (\forall \world \in \negE~.~\neg\textrm{PartialEval}(\pred, \world))
  \end{array}\)!
  $\pred'$ = Solve($\phi$)
  if ($\pred'$ $\neq$ UNSAT): return $\pred'$
  else: return UNSAT
!\(\begin{array}{@{}l}
  \textrm{\textbf{ParamHoles}} : \pred \rightarrow [?\param] \\
    \textrm{\textbf{PartialEval}} : \pred \times \world \rightarrow \pred \\
\end{array}\)!
\end{lstlisting}
\vspace{-2.0em}
\caption{\technique{-L1}}
\figlabel{l1Sketch}
\end{wrapfigure}
\technique{-L1} fills in values for blank constant parameters ($?\param$) in
a predicate ($\pred$), under
the assumption that there are no blank expressions or predicates in $\pred$.
The input is the
predicate, a set of positive examples on
which $\pred$ must produce true ($\posE$), and a set of negative examples on which
$\pred$ must produce false ($\negE$). The result of \technique{-L1}
is a new predicate where all blanks in the input are replaced
with constant values.

\technique{} uses Rosette and the Z3 SMT solver~\cite{rosette, z3} to solve constraints.
To do so, we translate the incomplete predicate and examples into SMT constraints
(\figref{l1Sketch}).
\technique{-L1} builds a formula ($\phi$) for every example,
which asserts that there exists some value for each blank parameter ($?\param$) in the
predicate, such that the predicate evaluates to true on a positive example (and
false on a negative example). Moreover, for each blank parameter, we ensure that we
chose the same value across all examples. The algorithm uses two
auxiliary functions: 1)~\texttt{ParamHoles} returns the
set of blank parameters in the predicate, and 2)~\texttt{PartialEval}
substitutes input values from the example into a predicate and simplifies
it as much as possible, using partial evaluation~\cite{partialEval}.
A solution to this system of constraints allows us to
replace blank parameters with values that are consistent with all
examples. If no solution exists, we return \texttt{UNSAT} (unsatisfiable).

\subsection{\technique{-L2} : Feature Synthesis}
\seclabel{L2}



\ldips{2} consumes a predicate ($\pred$) with blank expressions ($?\expr$)
and blank parameters ($?\param$) and produces a completed predicate.
(An incomplete predicate may occur in a user-written ASP, or may be generated
by \ldips{3} to decide on a specific action
transition in the ASP.) To complete the predicate, \ldips{2} also receives sets
of positive and negative examples ($\posE$ and $\negE$), on which the predicate
should evaluate to \texttt{true} and \texttt{false} respectively.
Since the predicate guards an action transition, each positive example corresponds
to a demonstration where the transition is taken, and each negative example
corresponds to a demonstration where it is not. Finally, \ldips{2} receives
a type environment ($\env$) of candidate expressions to plug into blank
expressions and a maximum depth ($\maxDepth$).
If \ldips{2} cannot complete the predicate to satisfy the examples,
it returns \texttt{UNSAT}.

The \ldips{2} algorithm (\figref{L2}) proceeds in several steps. 1)~It
enumerates a set of candidate expressions ($\exprs$) that do not exceed the
maximum depth and are dimension-constrained (line~\ref{line:enumFeatures}).
2)~It fills the blank expressions in the predicate using the candidate
expressions computed in the previous step, which produces a new predicate
$\pred'$ that only has blank parameters (line~\ref{line:fillExpressions}).
3)~It calls \ldips{1} to fill in the blank parameters and returns that
result if it succeeds. 4)~If \ldips{1} produces \texttt{UNSAT}, then
the algorithm returns to Step 2 and tries a new candidate expression.

\newcommand{\enumf}{\texttt{EnumFeatures}}

The algorithm uses the \enumf{} helper function to enumerate all expressions
up to the maximum depth that are type- and dimension- correct.
The only expressions that
can appear in predicates are scalars, thus the initial call to \enumf{} asks
for expressions of type $\dims$. (Recursive calls encounter other types.)
\enumf{} generates expressions by applying all possible operators to
 sub-expressions, where each sub-expression is itself produced by a
recursive call to \enumf{}.

The base case for the recursive definition is when
$\depth=0$: the result is the empty set of expressions. Calling
 \enumf{} with $\depth=1$ and type $T$ produces the subset
of input identifiers $\inputVar$ from the type environment $\env$ that have
the type $\type$. Calling \enumf{} with $\depth>1$
type $\type$ produces all expressions $\expr$, including those that involve
operators.
For example, if \enumf{} generates $\expr_1 \bOpE \expr_2$ at depth $\depth+1$,
it makes recursive calls to generates the expressions $\expr_1$ and $\expr_2$
at depth $\depth$. However, it ensures that the type and dimension of $\expr_1$
and $\expr_2$ are compatible with the binary operator $\bOpE$. For example,
if the binary operator is $+$, the sub-expressions must both be scalars or
vectors with the same dimensions. This type and dimension constraint allows us
to exclude a large number of meaningless expressions from the search space.
\figref{L2} presents a subset of the recursive rules of expansion for
\enumf.

Even with type and dimension constraints, the search space of
\enumf{} can be intractable. To further reduce the search space, the
function uses a variation of signatures equivalence~\cite{exampleTransit}, that
we extend to support dimensions. A naive approach to expression enumeration would
generate type- and dimension correct expressions that represent different
functions, but produce the same result on the set of examples.
For example, the expressions $|x|$ and $x$ represent different functions
with the same type and dimension. However, if our demonstrations only have positive
values for $x$, there is no reason to consider both expressions, because they
are equivalent given our demonstrations. We define the \emph{signature} ($\sig$) of
an expression as its result on the sequence of demonstrations, and we prune
expressions with duplicate signatures at each recursive call, using
the \texttt{SigFilter} function.

\begin{figure}[tbh]
\begin{minipage}[htb!]{0.42\textwidth}
  \scriptsize
\begin{lstlisting}[language=Python, mathescape=true,
  basicstyle=\scriptsize\ttfamily, numbers=none, numbers=left, escapechar=!]
!\(\begin{array}{@{}l}
  \textrm{\textbf{L2}} :
      \nat \times \env \times \worlds \times \worlds \times \pred
      \rightarrow \pred~||~\textrm{UNSAT}
\end{array}\)!
L2($\depth$, $\env$, $\posE$, $\negE$, $\pred$):
  $\exprs$ = EnumFeatures($\maxDepth, \env, \dims, \posE \cup \negE, ?\expr$) !\label{line:enumFeatures}!
  for $\pred'$ in FillExpressions($\exprs, \pred$); !\label{line:fillExpressions}!
    result = L1($\posE, \negE, \pred'$) !\label{line:result}!
    if (result $\neq$ UNSAT):
      return result !\label{line:return}!
  return UNSAT !\label{line:unsat}!
\end{lstlisting}
  \scriptsize
  \(\begin{array}{@{}l}
    \quad \textrm{\textbf{FillExpressions}} : \exprs \times \pred \rightarrow \{\pred\}
  \end{array}\)
  \(\begin{array}{@{}l}
    \quad \textrm{\textbf{SigFilter}} : \{ \langle \expr, \sig \rangle \} \rightarrow
                         \{ \langle \expr, \sig \rangle \} \\
  \end{array}\)
\end{minipage}
\vrule
\hspace{0.5em}
\begin{minipage}[htb!]{0.45\textwidth}
  \scriptsize
  \(\begin{array}{@{}l}
    \textrm{\textbf{EnumFeatures}} :
    \nat \times \env \times \type \times \worlds \times \expr
    \rightarrow \{\langle \expr, \sig \rangle\} \\
    \textrm{EnumFeatures}(0, \env, \type, \worldSet, \expr) = \{~\} \\
    \textrm{EnumFeatures}(\depth + 1, \env, \type, \worldSet, ?\expr)
    = \textrm{SigFilter(} \\
    \quad \{\textrm{EnumFeatures}(\depth, \env, \type, \worldSet, \expr),  \forall \expr : \type \in \env\}~\cup \\
    \quad \{\textrm{EnumFeatures}(\depth + 1, \env, \type, \worldSet, \uOp{?\expr}),
     \forall \uOpE : \type' \rightarrow \type \in \env\}) \\
    \textrm{EnumFeatures}(\depth+1, \env, \type, \worldSet, \const) =
    \{~\langle \const, \sig \rangle  \\
    \quad \given \sig = [\const,\dots,\const],
    \forall \world^i \in \worldSet \}\\
    \textrm{EnumFeatures}(\depth+1, \env, \type, \worldSet, \inputVar) =
    \{~\langle \inputVar, \sig \rangle\ \\
    \quad \given \sig = [\world^i.\inputVar,\dots,\world^n.\inputVar],
    \forall \world^i \in \worldSet \} \\
    \textrm{EnumFeatures}(\depth+1, \env, \type, \worldSet, \uOp{\expr}) =
    \{~\langle \uOp{\inputVar}, \sig \rangle \\
    \quad \given \uOpE : \type' \rightarrow \type \in \env,~\sig =
    [\interp(\uOp{\expr'},\world) \given \world \in \worldSet], \\
    \qquad\forall \expr' \in
    \textrm{EnumFeatures}(\depth, \env, \type', \worldSet, ?\expr) \} \\
  \end{array}\)
\end{minipage}
\caption{\technique-L2}
\figlabel{L2}
\end{figure}

\subsection{\technique{-L3} : Predicate Synthesis}
\seclabel{L3}
%
%
Given a set of demonstrations ($\demos$), \technique{-L3} returns a complete ASP
that is consistent with $\demos$.
The provided type environment
$\env$ is used to perform dimension-informed enumeration, up to a specified
maximum depth $\maxDepth$.
The \technique{-L3} algorithm (\figref{L3}) proceeds as follows.
\begin{paraenum}
  \item It separates the demonstrations into sub-problems consisting of
    action pairs, with positive and negative examples, according to the
    transitions in $\demos$.
  \item For each subproblem, it generates candidate predicates with
     maximum depth $\maxDepth$.
  \item For each candidate predicate, it invokes \technique{-L2}
    with the corresponding examples and the resulting expression, if one is
    returned, is used to the guard the transition for that sub-problem.
  \item If all sub-problems are solved, it composes them into an ASP ($p$).
\end{paraenum}

\begin{wrapfigure}[13]{r}{0.38\textwidth}
  \vspace{-1.7em}
\begin{lstlisting}[language=Python, mathescape=true, basicstyle=\scriptsize\ttfamily,
  numbers=none, escapechar=!]
!\(\begin{array}{@{}l}
  \textrm{\textbf{L3}} : \nat \times \demos \rightarrow \pol~||~\textrm{UNSAT}
\end{array}\)!
L3($\maxDepth$, $\demos$):
  $\partialPol$ = {}
  problems = DivideProblem($\demos$)
  for x $\in$ problems:
    Solution = False
    for $\pred\in$EnumPredicates($\maxDepth$):
      result = L2($\maxDepth, x.\posE, x.\negE, \pred$)
      if (result $\neq$ UNSAT):
        $\partialPol = \partialPol~\cup$result
        Solution = True
        break
  if (Solution):
    return MakeP(problems,$\partialPol$)
  else: return UNSAT
\end{lstlisting}
  \vspace{-1em}
  \caption{\technique{-L3}}
  \figlabel{L3}
\end{wrapfigure}

\technique{-L3} divides synthesis into sub-problems, using
the \texttt{DivideProblem} helper function, to address scalability.
\texttt{DivideProblem}
identifies all unique transitions from a starting action ($\startAction$) to
a final action ($\action_f$), and pairs of positive and negative examples
$\{\langle \posE^{s \rightarrow f}, \negE^{s \rightarrow f} \rangle\}$,
that demonstrate
transitions from $\startAction$ to $\action_f$, and transitions from $\startAction$
to any other final state respectively.
As an example sketch generated by \texttt{DivideProblems}, consider the
partial program shown in \cref{l3-incomplete-asp-example}.

Given the sketch generated by \texttt{DivideProblem}, \technique{-L3}
employs \texttt{EnumPredicates} to enumerate predicate structure.
\texttt{EnumPredicates} fills predicates holes $?\pred$ with
predicates $\pred$ according to the ASP grammar in \cref{syntax}, such that all expressions
$\expr$ are left as holes $?\expr$, and all constants $\thresh$ are left as
repairable parameter holes
$?\param$. Candidate predicates are enumerated in order of increasing
size until the maximum depth $\maxDepth$ is reached, or a solution is found.
For each candidate predicate $\pred$, and corresponding
example sets
$\posE$ and $\negE$, the problem reduces to one amenable to \technique{-L2}.
If a satisfying solution
for all $\pred$ is identified by invoking \technique{-L2},
they are composed into the policy $p$ using
\texttt{MakeP}, otherwise UNSAT is returned, indicating
that there is no policy consistent with the demonstrations.



\section{Evaluation}
  \seclabel{evaluation}
  We now present several experiments that evaluate
\begin{paraenum}
  \item the performance of ASPs synthesized by \technique{},
  \item the data-efficiency of \technique{}, compared to training an NN,
  \item the generalizability of synthesized ASPs to novel scenarios and
  \item the ability to repair ASPs developed in simulation, and to transfer them
  to real robots.
\end{paraenum}
Our experiments use three ASPs from two application domains.
\begin{paraenum}
\item From \emph{robot soccer}, the \textbf{attacker} plays the primary
offensive role, and use the fraction of scored goals over attempted goals as its
success rate.
\item From \emph{robot soccer}, the \textbf{deflector} executes one-touch
passes to the attacker, and we use the fraction successful passes over
attempted passes as its success rate.
\item From \emph{autonomous driving}, the \textbf{passer}  maneuvers through
slower traffic, and we use the fraction of completed
passes as its success rate.
\end{paraenum}
We use reference ASPs to build a dataset of demonstrations. For robot soccer,
we use ASPs that have been successful in RoboCup tournaments.
For autonomous driving, the reference ASP encodes
user preferences of desired driving behavior.

\subsection{Performance of Synthesized ASPs}
\seclabel{performance}

\begin{wrapfigure}[9]{r}{0.35\textwidth}
  \vspace{-1.65em}
  \begin{center}
    \begin{tabular}{|@{\;}l|@{\;}c|@{\;}c|@{\;}c| }
      \hline
      \multirow{2}{*}{\textbf{Policy}} &
      \multicolumn{3}{c|}{\textbf{Success Rates ($\%$)}}                                          \\
      \cline{2-4}
                                       & \textbf{Attacker} & \textbf{Deflector} & \textbf{Passer} \\
      \hline
      Ref                              & 89                & 86                 & 81              \\
      \hline
      LSTM                             & 78                & 70                 & 55              \\
      \hline
      NoDim                            & 78                & 76                 & 60              \\
      \hline
      L1                  & 75                & 85                 & 70              \\
      \hline
      L2                  & 89                & 80                 & 65              \\
      \hline
      L3                  & 87                & 81                 & 74              \\
      \hline
    \end{tabular}
  \vspace{-.6em}
  \caption{Success rates for different ASPs on three different
    behaviors in simulated trials.}
  \figlabel{simPerformance}
  \end{center}
\end{wrapfigure}
We use our demonstrations to 1)~train an LSTM that encodes the ASP, and
2)~synthesize ASPs using \technique{}-L1, \technique{}-L2, and \technique{}-L3.
For training and synthesis, the training set consists of 10, 20, and 20
trajectories for the attacker, deflector, and passer. For
evaluation, the test sets consists of 12000, 4800, and 4960 problems.
\figref{simPerformance} shows that  \technique{} outperforms
the LSTM in all cases.
For comparison, we also evaluate the reference ASPs, which can outperform the synthesized ASPs.
The best \technique{} ASP for deflector was within $1\%$ of the reference,
while the LSTM ASP was $16\%$ worse.

\subsection{Effects of Dimensional Analysis}
Dimensional analysis enables tractable synthesis of ASPs and improves
the performance of the learned policies. We evaluate the impact of dimensional
analysis by synthesizing policies with four variations
of \technique{-L3}, the full algorithm, a variant with only dimension based
pruning, with only signature-based pruning, and with
no expression pruning, all with a fixed depth of $3$. In \figref{simPerformance2} we report the number
of expressions enumerated for each variant, for each of our behaviors, as well
as the performance of each of the resulting policies.

\begin{wrapfigure}[7]{l}{0.5\textwidth}
  \vspace{-1em}
  \centering
  \begin{tabular}{ |@{\;}l|@{\;}c|@{\;}c|@{\;}c||@{\;}c|@{\;}c|@{\;}c| }
    \hline
    \multirow{2}{*}{\textbf{Policy}} &
    \multicolumn{3}{c|}{\textbf{\# Enumerated}} & \multicolumn{3}{c|}{\textbf{Success Rate \%}} \\
    \cline{2-7}
    & \textbf{Atk} & \textbf{Def} & \textbf{Pass} & \textbf{Atk} & \textbf{Def} & \textbf{Pass} \\
    \hline
    LDIPS-L3  & 175  & 174 & 345 & 87 & 81 & 74             \\
    \hline
    Dimension Pruning  & 696  & 696  & 1230 & 87 & 81 & 74             \\
    \hline
    Signature Pruning  & 4971 & 5013 & 366 & 78 & 76 & 60             \\
    \hline
    No Pruning & 14184 & 14232 & 7528 & - & - & -             \\
    \hline
  \end{tabular}
  \caption{Features enumerated at depth 3.}
  \figlabel{simPerformance2}
\end{wrapfigure}

For all of our behaviors, the number of expressions enumerated without dimensional
analysis or dimension informed signature pruning increases by orders of magnitude.
With this many possible candidate expressions, synthesis becomes intractable, and
as such, without pruning, we cannot synthesize a policy to evaluate at all.
Further, the performance of the ASPs synthesized with only signature pruning are
consistently
worse than \technique{-L3} and the difference is most stark in the passer ASP,
with a performance difference of $14\%$ between them.

\subsection{Data Efficiency}
\seclabel{dataEfficiency}

\technique{} can synthesize  ASPs with far fewer
demonstrations than the LSTM. To illustrate this phenomenon, we train
the LSTM with
\begin{paraenum}
  \item the full LSTM training demonstrations(\emph{LSTM-Full},
  \item half of the training demonstrations  (\emph{LSTM-Full}), and
  \item the demonstrations that \technique{} uses (\emph{LSTM-Synth}), which
  is a tiny fraction of the previous two training sets.
\end{paraenum}
\begin{wrapfigure}[4]{r}{0.41\textwidth}
  \vspace{-0.5em}
  \begin{center}
    \begin{tabular}{ |@{\;}l|@{\;}c|@{\;}c|@{\;}c|@{\;}c|}
      \hline
      \multirow{2}{*}{\textbf{Policy}} &
      \multicolumn{2}{c|}{\textbf{Attacker}}
                                       & \multicolumn{2}{c|}{\textbf{Deflector}}                            \\
      \cline{2-5}
                                       & \textbf{($\%$)}                         & \textbf{N}
                                       & \textbf{($\%$)}                         & \textbf{N}               \\
      \hline
      LSTM-Full                        & 78                                      & 778408     & 70 & 440385 \\
      \hline
      LSTM-Half                        & 32                                      & 389204     & 61 & 220192 \\
      \hline
      LSTM-Synth                       & 25                                      & 750        & 38 & 750    \\
      \hline
      \technique{}                     & 87                                      & 750        & 81 & 750    \\
      \hline
    \end{tabular}
  \vspace{-0.5em}
    \caption{Performance \vs{} \# of examples ($N$).}
  \figlabel{dataEfficiency}
  \end{center}
\end{wrapfigure}
\figref{dataEfficiency} shows how the performance of the LSTM degrades
as we cut the size of the training demonstrations. In particular, when the LSTM
and \technique{} use the same training demonstrations, the LSTM fares significantly
worse ($57\%, 47\%$ inferior performance).

\subsection{Ability to Generalize From Demonstrations}

\begin{wrapfigure}{r}{0.43\textwidth}
\vspace{-1em}
\centering
\subcaptionbox{Reference ASP}{
  \includegraphics[width=0.2\textwidth]{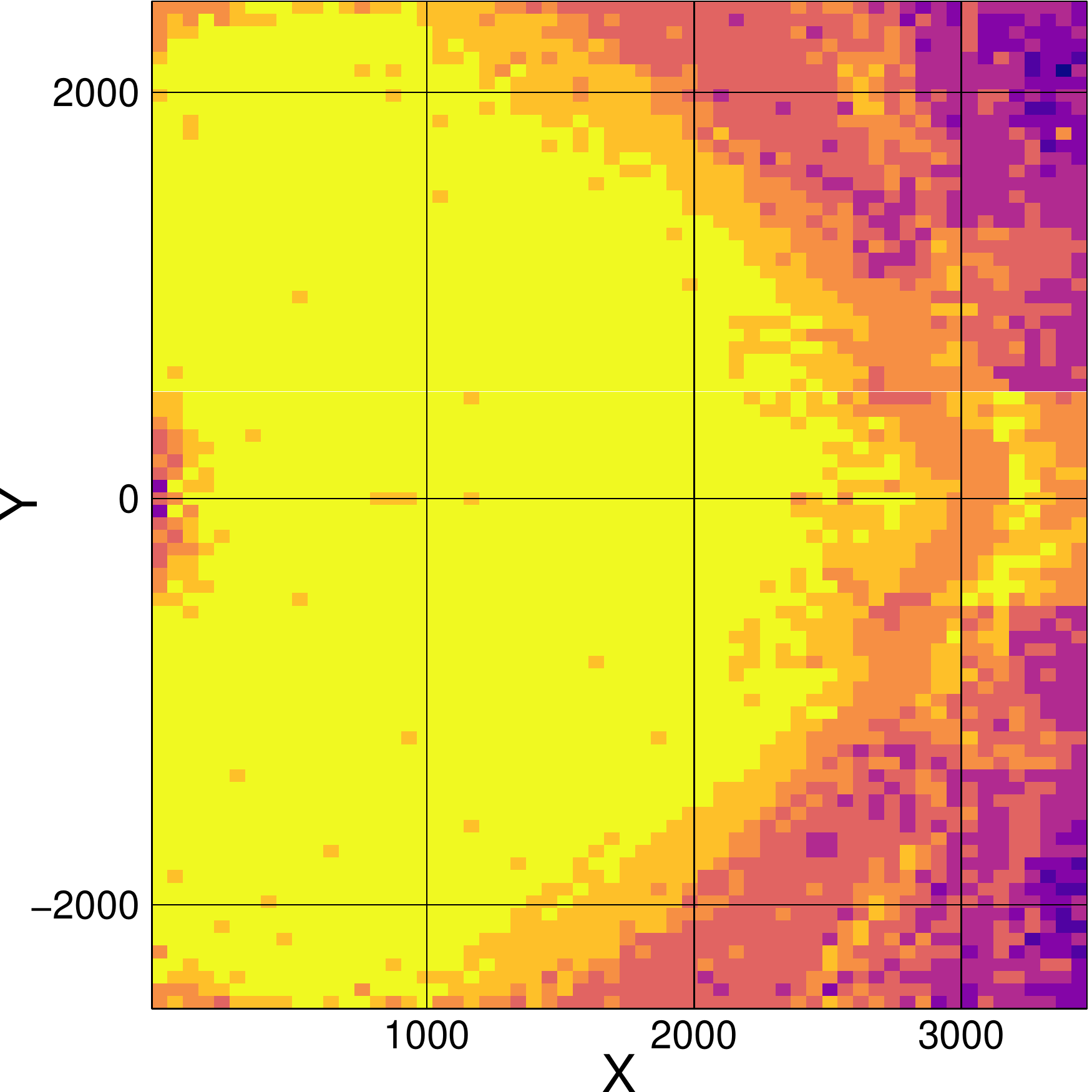}}\hspace{0.5em}%
\subcaptionbox{\technique{}-L3 ASP}{
  \includegraphics[width=0.2\textwidth]{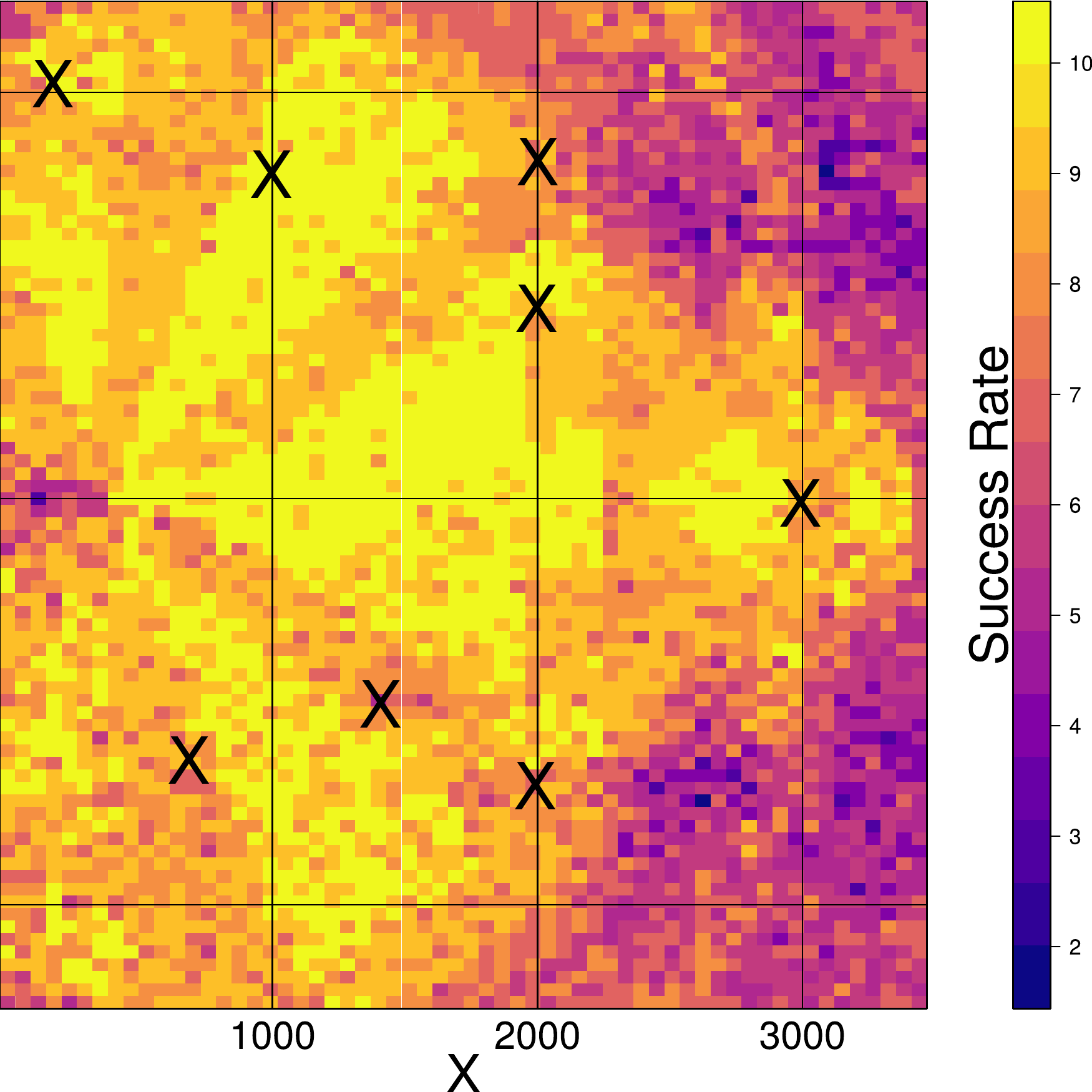}}%
\caption{Attacker success rate with varying ball positions. Training locations
are marked with an X.}
\figlabel{heatmap}
\vspace{-1em}
\end{wrapfigure}
A key requirement for an LfD algorithm is its ability to generalize to novel
problems. This experiment shows that an \emph{attacker} ASP, synthesized
using \technique{}-L3 and only ten demonstrations, can score a goal when
a moving ball is placed at almost any reasonable position on the field.
On each run, the attacker starts at the origin (\cref{fig:heatmap}). We
discretize the soccer field, place the ball at a discrete point, and
set the ball in motion in $10$ possible directions (12,000 total runs). Thus,
each point of the heatmap shows the attacker's success rate
on all runs that start at that point.
The figure shows the performance of the \technique{}-L3 synthesized ASP
on ten demonstration runs that start from the eight marked positions. The
synthesized ASP generalizes to problems that are significantly different
from the training examples. Moreover, its performance degrades on exactly
the same region of the field as the reference ASP (i.e., when the ball
is too far away for the attacker to intercept).

\subsection{Transfer From Sim To Real}
\begin{wrapfigure}[7]{r}{0.4\textwidth}
  \begin{centering}
  \scriptsize
    \vspace{-1.5em}
    \setlength\tabcolsep{3pt} 
    \begin{tabular}{@{}|@{\;}l|@{\;}c|@{\;}c|@{\;}c|@{\;}c|@{\;}c|@{\;}c|}
      \hline
      \multirow{2}{*}{\textbf{Scenario}}
        & \multicolumn{3}{c|}{\textbf{Attacker}}
        & \multicolumn{3}{c|}{\textbf{Deflector}} \\
      \cline{2-7}
        & \textbf{Ref} & \textbf{LSTM} & \textbf{L3} & \textbf{Ref} & \textbf{LSTM} & \textbf{L3} \\
      \hline
      Sim & 89 & 78 & 87 & 86 & 70 & 81 \\
      \hline
      Real & 42 & 48 & 50 & 70 & 16 & 52 \\
      \hline
      Repaired & 70 & - & 64 & 78 & -  & 72 \\
      \hline
    \end{tabular}
  \caption{Sim-to-real performance, and ASP repair. }
  \figlabel{realPerformance}
  \end{centering}
\end{wrapfigure}

ASPs designed and tested in simulation frequently suffer from degraded
performance when run on real robots. If the ASP is hand-written and includes
parameters it may be repaired by parameter optimization, but NN ASPs are much
harder to repair without significant additional data collection and retraining.
However,
\technique{} can make the sim-to-real transfer process significantly easier. For this experiment, using the
attacker and deflector, we 1)~synthesize ASPs in a simulator, and 2)~deploy them
on
a real robot. Predictably, the real robot sees significantly degraded
performance on the reference ASP, the learned LSTM ASP, and the
LDIPS-synthesized ASP. We use a small variant of \ldips{1} (inspired
by SRTR~\cite{holtz2018interactive}) on the reference and LDIPS ASPs:
to every parameter ($x$) we add a blank adjustment ($x+?x$), and synthesize
a minimal value for each blank, using ten real-world demonstration runs.
The resulting ASPs perform significantly better, and are much
closer to their performance in the simulator (\cref{fig:realPerformance}).
This procedure is ineffective on the LSTM: merely ten demonstration runs
have minimal effect on the LSTMs parameters. Morever, gathering a large
volume of real-world demonstrations is often impractical.

\section{Conclusion}
  \seclabel{conclusion}
In this work, we presented an approach for learning action selection policies for
robot behaviors utilizing layered dimension informed program synthesis (\technique{}).
This work
composes skills into high-level behaviors using a small number of demonstrations
as human-readable programs. We demonstrated
that our technique generates high-performing policies with
respect to human-engineered and learned policies in two different domains.
Further, we showed that these policies could be transferred from
simulation to real robots by utilizing parameter repair.




\section*{Acknowledgments}
\vspace{-1em}
This work was partially supported by the National Science Foundation under
grants CCF-2102291 
and CCF-2006404, and by JPMorgan Chase \& Co. 
In addition, we acknowledge support from Northrop Grumman Mission Systems’
University Research Program. Any views or opinions expressed herein are solely those of the authors listed.

\clearpage
\bibliography{references}

\end{document}